\documentclass[%
 reprint,
 amsmath,amssymb,
 aps,
]{revtex4-2}

\usepackage{graphicx}% Include figure files
\usepackage{dcolumn}% Align table columns on decimal point
\usepackage{bm}% bold math
\usepackage{comment}

\usepackage{xcolor}

\begin{document}

\preprint{APS/123-QED}

\title{Hardware-Friendly Implementation of Physical Reservoir Computing with CMOS-based Time-domain Analog Spiking Neurons}% Force line breaks with \\

\author{Nanako Kimura}
 \altaffiliation{Department of Electrical Engineering and Information Systems, the University of Tokyo}%Lines break automatically or can be forced with \\
 \altaffiliation{Equally contributed authors.}%Lines break automatically or can be forced with \\
\author{Ckristian Duran}
 \altaffiliation{Systems Design Lab., School of Engineering, the University of Tokyo}%Lines 
 \altaffiliation{Equally contributed authors.}%Lines break automatically or can be forced with \\
\author{Zolboo Byambadorj}
 \altaffiliation{Systems Design Lab., School of Engineering, the University of Tokyo}%Lines 
\author{Ryosho Nakane}
 \altaffiliation{Systems Design Lab., School of Engineering, the University of Tokyo}%Lines 
 \altaffiliation{Also at Department of Electrical Engineering and Information Systems, the University of Tokyo}%
 \altaffiliation{Tetsuya Iizuka and Ryosho Nakane are corresponding authors.}%Lines break automatically or can be forced with \\
\author{Tetsuya Iizuka}
 \altaffiliation{Systems Design Lab., School of Engineering, the University of Tokyo}%Lines 
 \altaffiliation{Also at Department of Electrical Engineering and Information Systems, the University of Tokyo}%
 \altaffiliation{Tetsuya Iizuka and Ryosho Nakane are corresponding authors.}%Lines break automatically or can be forced with \\

\date{\today}% It is always \today, today,

\begin{abstract}
This paper introduces an analog spiking neuron that utilizes a time-domain information, i.e., a time interval of two signal transitions and a pulse width, to construct a spiking neural network (SNN) for
a hardware friendly physical reservoir computing (RC) on a complementary metal-oxide-semiconductor (CMOS) platform. A neuron with leaky integrate-and-fire is realized by employing two voltage-controlled oscillators (VCOs) with opposite sensitivities to the internal control voltage, and the neuron connection structure is restricted by use of only 4 neighboring neurons on the 2-dimensional plane to feasibly construct a regular network topology. Such a system enables us to compose a SNN with a counter-based readout circuit, which simplifies the hardware implementation of the SNN. Moreover, another technical advantage thanks to the bottom-up integration is the capability of dynamically capturing every neuron state in the network, which can significantly contribute to finding guidelines on how to enhance the performance for various computational tasks in temporal information processing. Diverse nonlinear physical dynamics needed for RC can be realized by collective behavior through dynamic interaction between neurons, like coupled oscillators, despite the simple network structure. With behavioral system-level simulations, we demonstrate physical RC through short-term memory and exclusive OR tasks, and the spoken digit recognition task with an accuracy of 97.7\,\% as well. Our system is considerably feasible for practical applications and also can be a useful platform to study the mechanism of physical RC.  
\end{abstract}

\maketitle

%\tableofcontents

\section{\label{sec:level1}Introduction}
Spiking neural networks (SNNs) utilizing electrical circuits have attracted great interest from a wide range of research areas, and they are rapidly progressing to form an interdisciplinary research area that can be roughly divided into two groups: neuroscience and machine-learning computing. The former group aims at clarifying bio-related phenomena, including those in human brains, by exploring how a large-size neural network behaves when electrical spiking neurons and synapses are connected in a bottom-up fashion. The latter group aims at developing energy-efficient information processing technologies that have extremely lower-power consumption than software-implemented feed-forward Deep Neural Networks (DNNs) with backpropagation learning algorithms. This is because, if the demand for such DNNs technology keeps growing, the energy consumption is becoming of critical concern, and in the worst-case scenario~\cite{DEVRIES20232191}, Google’s AI alone could consume as much electricity as a country such as Ireland. So far, various types of SNN systems have been implemented with circuits~\cite{merolla2014million, furber2014spinnaker, pnas1604850113, pfeil2013six, moriya2022fully, bofill2003learning, benjamin2014neurogrid, azghadi2015programmable, qiao2015reconfigurable, mitra2008real, chen2023cmos}.

Whereas SNN-based systems with digital circuits have advantages, such as robustness to the environmental noise and direct benefits from the Complementary Metal Oxide Semiconductor (CMOS) technology scaling, SNN-based systems with analog CMOS circuits can bring about more energy- and area-efficient features of the computing system~\cite{joubert2012hardware}. 
To benefit from these properties, there are a lot of studies designing the spiking neuron circuit in an analog manner, especially with CMOS technology~\cite{wijekoon2008compact, linares1991cmos, mahowald1991silicon, wu2015cmos, hasan2020low, tamura2019izhikevich, yang2020analog, farquhar2005bio, wang2015compact}. 
However, the analog implementation is often more sensitive to the noise than its digital counterpart~\cite{soriano2014delay}.

To further advance the SNN technology with analog CMOS circuits, keys are the hardware feasibility and machine-learning computing framework including learning algorithm. For energy-efficient information processing of time-series data, one of the most attractive frameworks is reservoir computing (RC), in which a special type of recurrent neural networks (RNNs) is used as a “reservoir” for temporal nonlinear transformation of input data under fading memory influence. The feasibility of the RC system comes from the fact that the weights of the random/sparse neuron connections in the reservoir are fixed and those in the linear learner in the readout are adjustable. Hence, the capacity of rewritable memory storage is extremely smaller than that required for typical SNN-based systems. Furthermore, since the training algorithm for the optimization is a linear regression, i.e., an iterative optimization is unnecessary, it offers highly-adaptive, real-time, and energy-efficient computing through an on-line update of weights in a short time, as was demonstrated using a software-implemented RC ``echo-state network (ESN)''~\cite{vlachas2020backpropagation}. In addition to such features, the prominent feasibility is that nonlinear physical dynamics with fading memory can be used as a reservoir. This physics-oriented RC, called “physical RC”, motivates many researchers to create on-chip reservoir devices using various physical systems, e.g., 
analog circuits~\cite{appeltant2011information, zhao2016novel}, optics devices~\cite{Bueno:17, duport2012all}, optoelectronic devices~\cite{paquot2012optoelectronic, martinenghi2012photonic}, spin-tronic devices~\cite{kanao2019reservoir,nakane2021spin}, FeFET devices~\cite{toprasertpong2022reservoir}, memristors~\cite{du2017reservoir, moon2019temporal, zhong2022memristor}, water~\cite{fernando2003pattern}, and soft materials~\cite{nakajima2015soft}, etc.
Indeed some of those devices achieve high computing performances, but the most of them require high-speed analog-to-digital converters (ADCs), leading to high power consumption and a large overhead of the system. Moreover, the entire system design for practical implementation onto a CMOS platform has not been established, more precisely, even system design guidelines for scale-up integration are not clear. This is partially because a major part of reservoirs thus far is a single device or network devices fabricated in a top-down manner, i.e., the fine adjustment through a bottom-up system construction has not been deeply explored. Thus, studying RC performance using various bottom-up-constructed neural networks can find system design guidelines, which considerably contribute to physical RC toward practical use. For this purpose, step-by-step clarification is infinitely preferable while using a simple network topology with increasing system size. On the other hand, some physical RC systems have been implemented using fully analog circuits, e.g., one reproduces Mackey-Glass dynamics using a feedback loop and ADCs~\cite{appeltant2011information}, whereas another faithfully reproduces a large-size network topology with a feedback loop through some recursive neuron connections, called ``cycle reservoir''~\cite{liang2022rotating}. Those systems achieve high computing performances and promising candidates for large-size RC systems, however, there remain issues in terms of area- and energy-efficiencies.

We have proposed CMOS-based analog modules for a spiking neuron with Leaky Integrate-and-Fire (LIF), a synapse with fading memory, and a signal weighting for a neuron connection, and studied time-domain SNNs where a time-series transmission signal inside the network has information expressed by the frequency of pulses produced by neuron firings as well as the width of a pulse from a synapse~\cite{chen2023cmos}. Those modules allow to construct area- and energy-efficient SNN systems since they can exclude the uses of operational amps and continuous-time or clocked comparators. After demonstrating the principal operations of implemented modules, we performed a simple task using physical RC while randomly connecting the modules to construct a reservoir on software. 

In this paper, we study on-chip physical RC systems using our CMOS-based time-domain analog modules~\cite{chen2023cmos} to 
realize a hardware-friendly implementation of the SNN reservoir in a bottom-up fashion. Since our neurons transmit and receive the analog information expressed by the frequency and the width of the pulses, which are 0/1 logic waveforms, in principle, our physical RC system has similar robustness to the noise as in the digital implementations. As a further improvement toward implementation, a simple counter module is also proposed to avoid the uses of power- and area-consuming ADCs. It is widely recognized that the random and small-world network topologies that are frequently used for neural networks~\cite{watts1998collective} are not feasible for the 2-dimensional chip implementation, since the lengths of many wirings between neurons can be unacceptably long over the chip area and the circuit design becomes too complicated to realize the scalability of the system. To exclude this difficulty, our SNN reservoir uses a regular network topology that has neuron connections only between 4 neighboring neurons in the plane. This network topology is extremely hardware-friendly, but it may have inferior capability of rich nonlinear dynamics indispensable for RC~\cite{kawai2019small}. It is worth noting that our spiking neuron circuit has characteristic dynamics originating from oscillation with fading memory. Hence, we expect that it can exhibit diverse nonlinear dynamics through the interactions with other modules in a network structure, like coupled spin-torque oscillators used for machine-learning computing~\cite{romera2018vowel}. Our challenge is to demonstrate that the network dynamics resulting from multiple neurons and their connections can overcome the disadvantage of the regular topology, which is very useful knowledge to wide-range studies aiming at the implementation of a large network-type reservoir on a chip.  
This paper clearly demonstrates the scalability of the proposed physical RC system with simulation results using different number of neurons.

\section{Circuit Architectures} \label{sec:circuit}

\begin{figure}[tb]
\includegraphics[width=\linewidth]{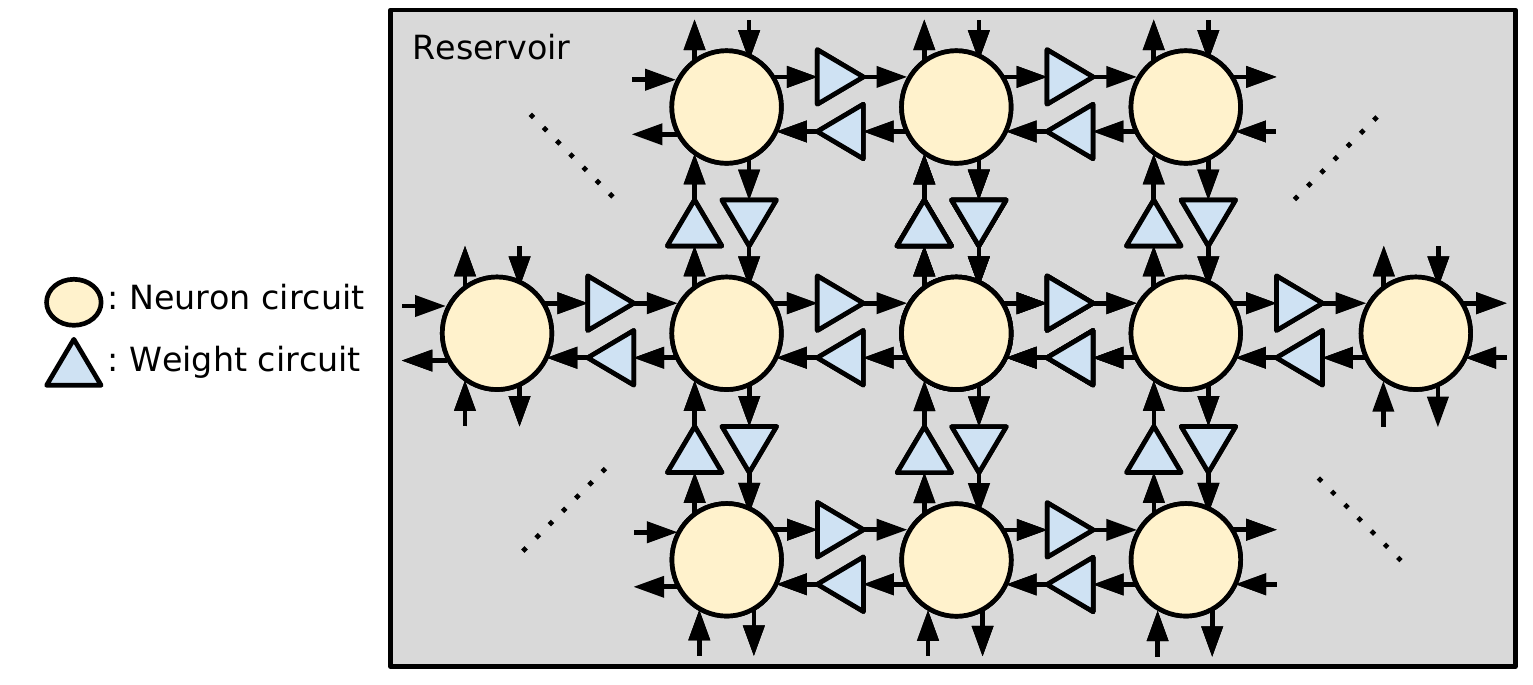}
\caption{\label{fig:inside_chip} On-chip reservoir network architecture in this paper, which consists of neuron circuits and weight circuits. 
The connections among the neurons in the reservoir are limited to only 4 neighboring neurons to reduce the complexity of the wiring in actual chip implementations.}
\end{figure}

\begin{figure}[tb]
\includegraphics[width=\linewidth]{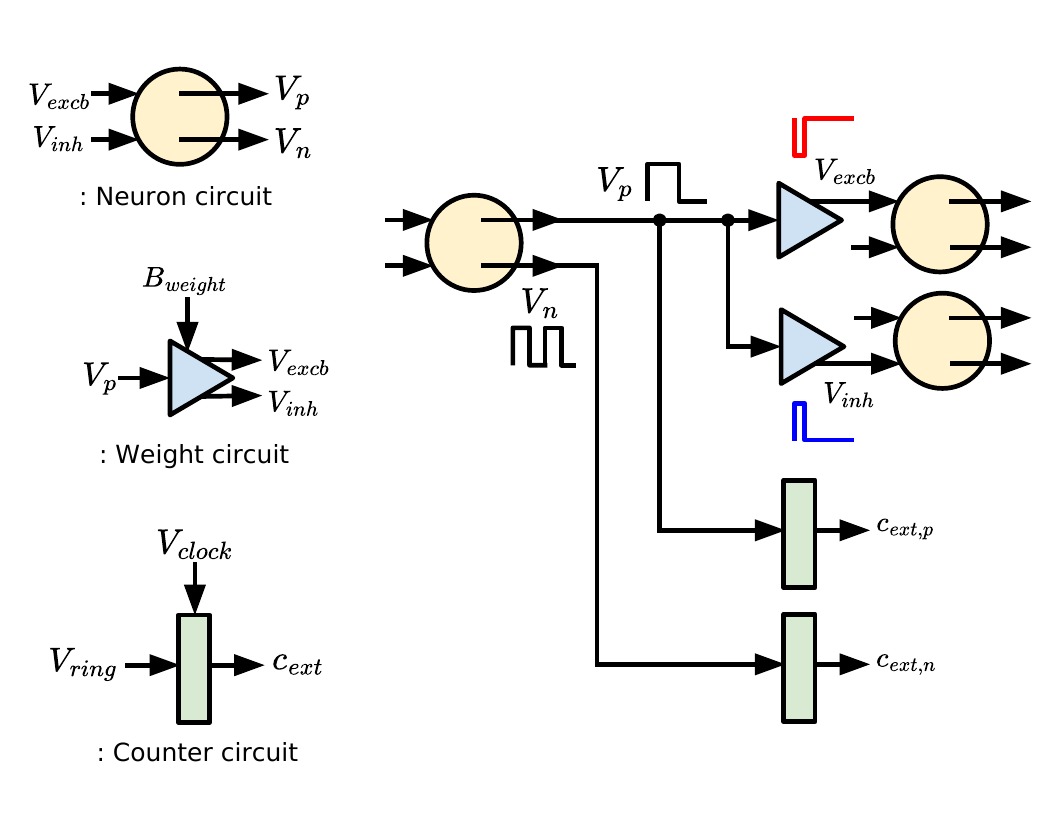}
\caption{\label{fig:chip_relation} Relationship between 3 circuits: neuron circuits, weight circuits, and counter circuits.
The pulse repetition rate (or frequency) of $V_{p}$ has a positive dependence on the internal control voltage of the neuron, while that of $V_{n}$ has a negative dependence. The pulse repetition rates of both pulses are detected by the counters to predict the internal state of the neuron, while only $V_{p}$ is also connected to the neighboring neurons through the weight circuits.
The weight circuit changes the pulse width of $V_{p}$ depending on the digital weight control word to generate either excitatory pulse $V_{excb}$ or inhibitory pulse $V_{inh}$ with specified width to be fed to the next neuron.}
\end{figure}

\begin{figure*}[t]
\includegraphics[width=\linewidth]{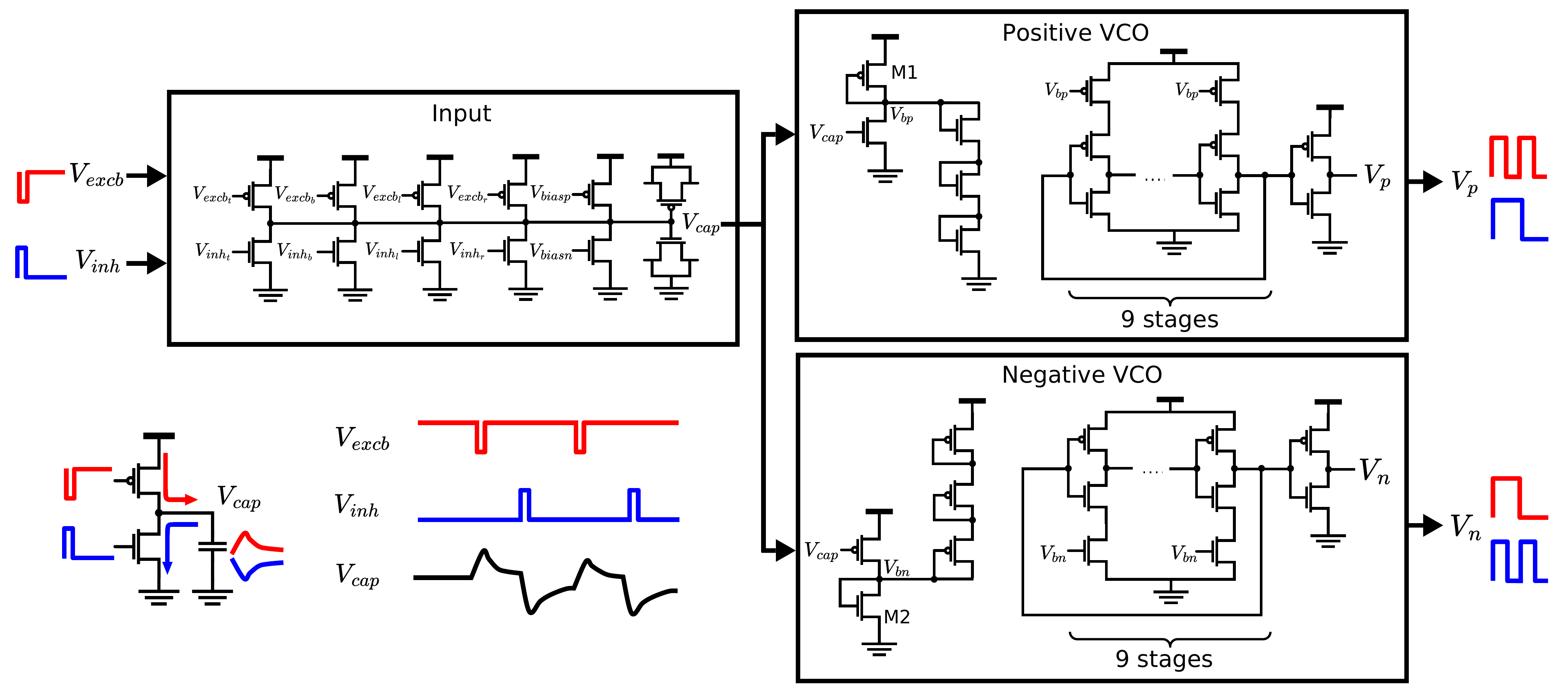}
\caption{\label{fig:neuron} Schematic diagram of the neuron circuit, which is composed of the input, positive VCO, and negative VCO blocks.
The neuron accepts two types of pulse inputs: excitatory pulse $V_{excb}$ and inhibitory pulse $V_{inh}$. The input block has 4 PMOS and 4 NMOS transistors to accept the excitation and inhibition pulses from 4 neighboring neurons. $V_{biasp}$ and $V_{biasn}$ are external bias voltages to tune the steady-state value of the internal control voltage $V_{cap}$.
The ring oscillator circuit in each VCO is composed of 9-stage inverters. Its oscillation frequency is controlled by PMOS current source with the control voltage $V_{bp}$ in the positive VCO, while by NMOS current source with $V_{bn}$ in the negative side. Dedicated biasing circuits that generate $V_{bp}$ and $V_{bn}$ are implemented together with the ring oscillators. The steady-state values of $V_{bp}$ and $V_{bn}$ are set by the diode-connected FETs in the biasing circuits. Both $V_{bp}$ and $V_{bn}$ decrease when $V_{cap}$ increases. In the positive VCO, this leads to the increase of its oscillation frequency, while in the negative one, the frequency decreases.}
\end{figure*}

\subsection{Reservoir Network}
The on-chip reservoir network in this paper that consists of spiking neurons and weighting modules is illustrated in Figure~\ref{fig:inside_chip}. As mentioned in Sect.~\ref{sec:level1}, the connections among the neurons in the reservoir are limited to only 4 neighboring neurons to reduce the complexity of the wiring in actual chip implementations, and simple counters are used for reading out the internal states of the neurons as shown in Figure~\ref{fig:chip_relation}.
As will be detailed in the next subsection, the neuron circuit includes two VCOs that generate two repetitive pulse outputs as spikes, $V_{p}$ and $V_{n}$. 
Here in our time-domain neural network, the neuron activity is expressed with the pulse repetition rate, while the connection weight is expressed with the pulse width. The next neuron changes its internal control voltage depending on the pulse repetition rate and width. The details of each building block and its schematic diagram will be discussed in the following subsections.

\subsubsection{Spiking Neuron Circuit}
The spiking neuron employed in this paper is based on the Leaky Integrate-and-Fire~(LIF) model originally proposed by Lapicque~\cite{abbott1999lapicque}.
As illustrated in Figure~\ref{fig:neuron}, our spiking neuron is composed of 3 blocks: input circuit, positive VCO, and negative VCO. 
$V_{excb}$ is a negative narrow pulse and is fed to one of the PMOS FETs to turn ON the PMOS for short period of time, which in turn slightly charges up the PMOS/NMOS capacitor to increase $V_{cap}$. In contrast, $V_{inh}$ is a positive narrow pulse and is fed to one of the NMOS FETs, then turns ON the NMOS for a short period to slightly discharge the capacitor to decrease $V_{cap}$. Therefore, $V_{cap}$ is tuned depending on the activities and weights of the neighboring neurons. When the input block does not receive any excitation and inhibition pulses, depending on the sub-threshold and gate leakage currents, $V_{cap}$ tends to go back to its steady-state voltage. In total, the leaky integration function is realized with this input block.

\begin{figure}[tb]
\includegraphics[width=0.7\linewidth]{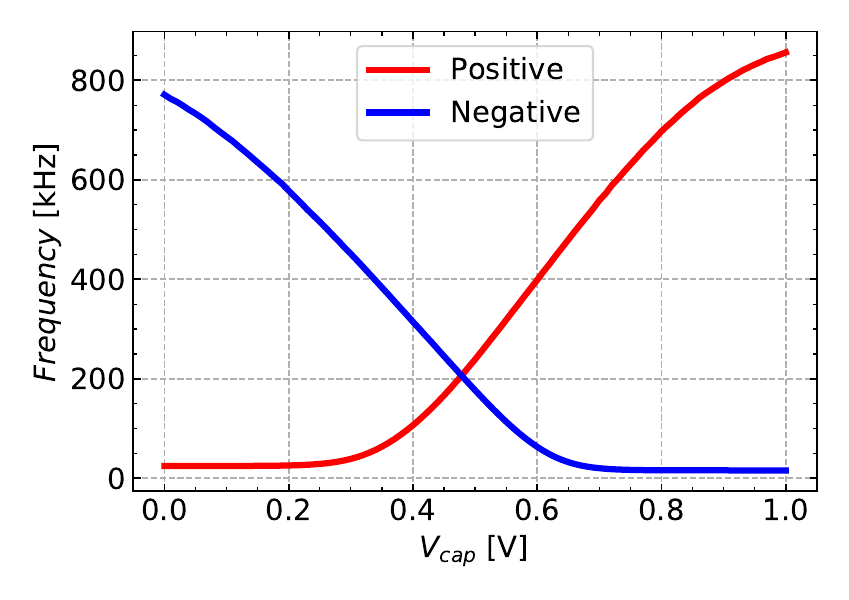}
\caption{\label{fig:freq} Frequencies of the positive and negative VCO outputs depending on the control voltage $V_{cap}$. Two VCOs have opposite dependence on $V_{cap}$. When $V_{cap}$ is small, the positive VCO becomes less sensitive to $V_{cap}$, while the negative VCO becomes less sensitive when $V_{cap}$ is large. To compensate for these insensitive regions, two VCOs are integrated in our neuron circuit.}
\end{figure}

Then, $V_{cap}$ is fed to both the positive and negative VCOs as illustrated in Figure~\ref{fig:neuron}. 
Figure~\ref{fig:freq} plots the simulation results of the positive and negative VCO output frequencies depending on the control voltage $V_{cap}$. 

\begin{figure}[tb]
\includegraphics[width=\linewidth]{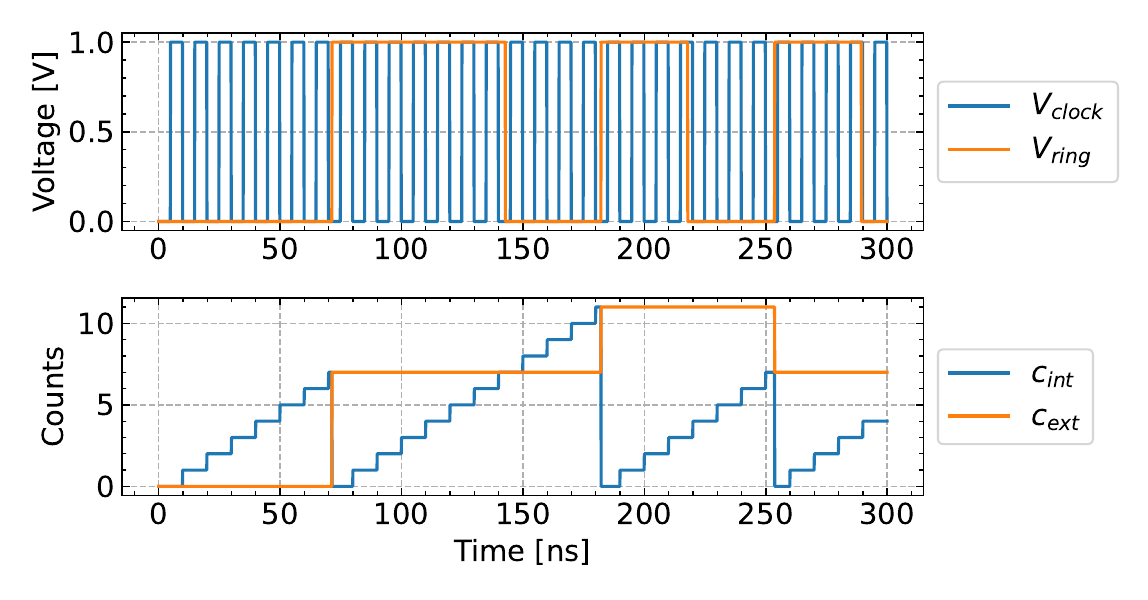}
\caption{\label{fig:counter_diagram} Waveforms of a counter circuit. The number of the rising edges of $V_{clock}$ is counted by the internal counter $c_{int}$. When $V_{ring}$ rises, the output $c_{ext}$ is set to $c_{int}$, while the internal counter value $c_{int}$ is reset to 0.}
\end{figure}

To predict $V_{cap}$ in a hardware-friendly manner, we use counters that is simply composed of a set of DFFs. Here to measure the frequencies or periods of the VCO outputs, the number of system clock periods, which is much shorter than the VCO output periods, is counted within the two rise transitions of the VCO outputs. Figure~\ref{fig:counter_diagram} shows the simulated waveforms of the counter circuit. The number of rising edges of system clock $V_{clock}$, which is 100\,MHz in this paper, is counted as $c_{int}$, then at the rise transition of the VCO output $V_{ring}$, the value of $c_{int}$ is latched to be read out as $c_{ext}$. At the same time, $c_{int}$ is reset to 0 to restart the count. Since the count value to be read out is updated at the rate of the VCO outputs, when the frequency of the VCO output is slow, i.e., the period is long, the counter value is rarely updated, which leads to the prediction error of $V_{cap}$. Therefore, we are using two VCOs with opposite dependence on $V_{cap}$ to maintain a sufficient update rate of the counter within the entire operation range. With this technique, we can successfully avoid using a dedicated ADC to predict the $V_{cap}$ without sacrificing the performance too much.

\begin{figure*}[tb]
\includegraphics[width=0.8\linewidth]{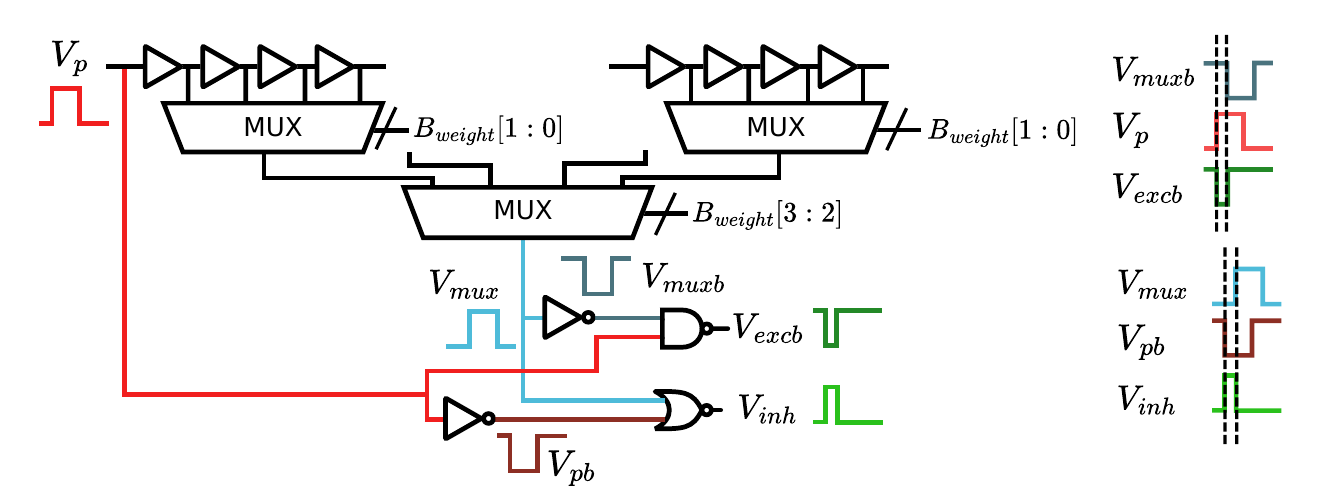}
\caption{\label{fig:weight} Schematic diagram of the weighting circuit, which was originally proposed in~\cite{chen2023cmos}. According to the $4$-bit digital tuning word $B_{weight}$, the multiplexer chooses a corresponding tap from the delay line composed of a series connection of buffers. The first two bits $B_{weight}[1:0]$ is for the first stage multiplexer and  the latter two bits  $B_{weight}[3:2]$ is for the second stage multiplexer. With the original and delayed input signals through NAND or NOR logic operation, the weighting circuit generates the pulse whose width is tuned depending on $B_{weight}$ as outlined by the waveforms at the right. Depending on whether the weight module is used for an excitatory or an inhibitory connection, either the output $V_{excb}$ or $V_{inh}$ is used for the next neuron.}
\end{figure*}

\subsubsection{Weighting circuit}
Figure~\ref{fig:weight} shows a schematic diagram of a weighting circuit.
If the pulse is wider, it charges or discharges $V_{cap}$ more in the input block of the next-stage neuron as shown in Figure~\ref{fig:neuron} to represent the strength or the weight of the connection between neurons.

\subsection{Construction of Behavioral Model}
To simulate the entire SNN system, behavioral models of the neuron and weighting circuits explained in the previous subsections are constructed, since simulating the entire SNN with a circuit simulator is unrealistically time consuming. The behavioral models are written in a Python environment based on the actual circuit behavior investigated with a standard analog IC simulator (Cadence HSPICE).

\begin{figure}[tb]
\includegraphics[width=0.9\linewidth]{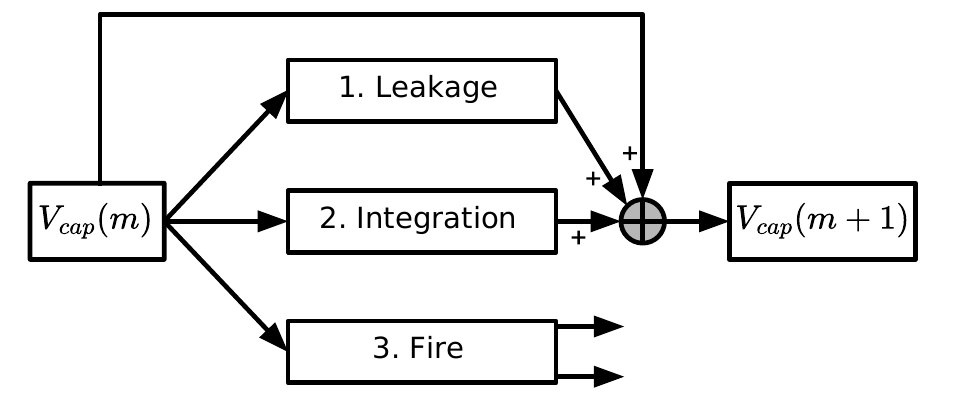}
\caption{\label{fig:code} Data flow of behavioral model simulation in one time step. The behavior of the neuron circuit is divided into three parts: Leakage, Integration, and Fire.}
\end{figure}

\begin{figure*}[tb]
\includegraphics[width=\linewidth]{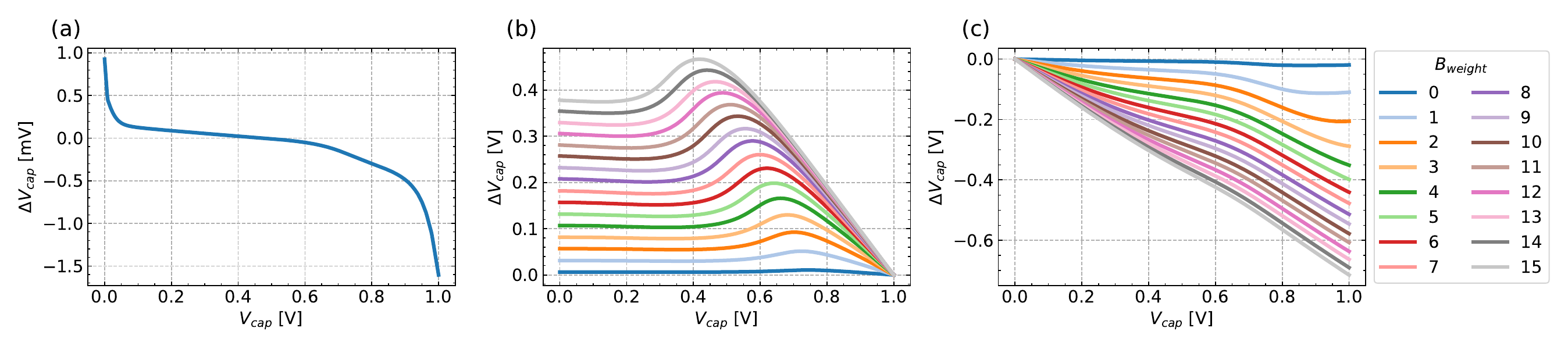}
\caption{\label{fig:leak&pulse} 
(a)Leakage feature of the neuron circuit. The change $\Delta V_{cap}$ due to the leakage in 10\,ns depends on the original $V_{cap}$ value. Since the steady-state value of $V_{cap}$ is designed to be about 0.5\,V, where $\Delta V_{cap}$ is almost 0, $\Delta V_{cap}>0$ for $V_{cap}<0.5$ while $\Delta V_{cap}<0$ for $V_{cap}>0.5$ due to the leakage.
Integration feature of the neuron circuit (b) for excitatory pulse input and (c) for inhibitory pulse input depending on the weight tuning word $B_{weight}$. The more voltage change $\Delta V_{cap}$ is observed with the larger weight $B_{weight}$ in both cases.}
\end{figure*}

As illustrated in Figure~\ref{fig:code}, we divide the behavior of the neuron circuit into three parts: Leakage, Integration, and Fire, whose following features are extracted.
\begin{enumerate}
  \item Leakage: The change of $V_{cap}$ ($\Delta V_{cap} = V_{cap}(m+1) - V_{cap}(m)$, where $m$ is the index of the simulation step) in a time step of behavioral simulation $10$\,ns due to the leakage depending on the original $V_{cap}$ value. Figure~\ref{fig:leak&pulse}(a) shows the circuit simulation result of $\Delta V_{cap}$ versus the original $V_{cap}$.
  \item Integration: The change of $V_{cap}$ with the pulse $V_{excb}$ or $V_{inh}$ applied.
  Figures~\ref{fig:leak&pulse}(b) and (c) show the circuit simulation result of $\Delta V_{cap}$ versus the original $V_{cap}$ depending on different $B_{weight}$ values.
  \item Fire: The frequencies of the outputs $V_{p}$ and $V_{n}$ versus $V_{cap}$. Figure~\ref{fig:freq} shows the circuit simulation results.
\end{enumerate}

\begin{figure}[tb]
\includegraphics[width=0.6\linewidth]{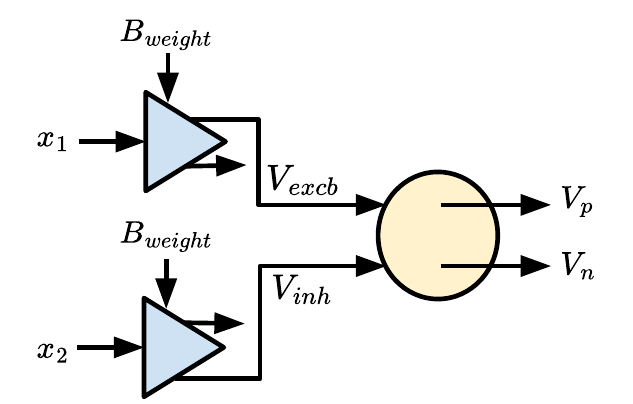}
\caption{\label{fig:behavior_test} Testbench circuit for comparison that consists of one neuron and two weighting circuits with an excitatory pulse $x_1$ and an inhibitory pulse $x_2$.} 
\end{figure}

\begin{figure}[tb]
\includegraphics[width=\linewidth]{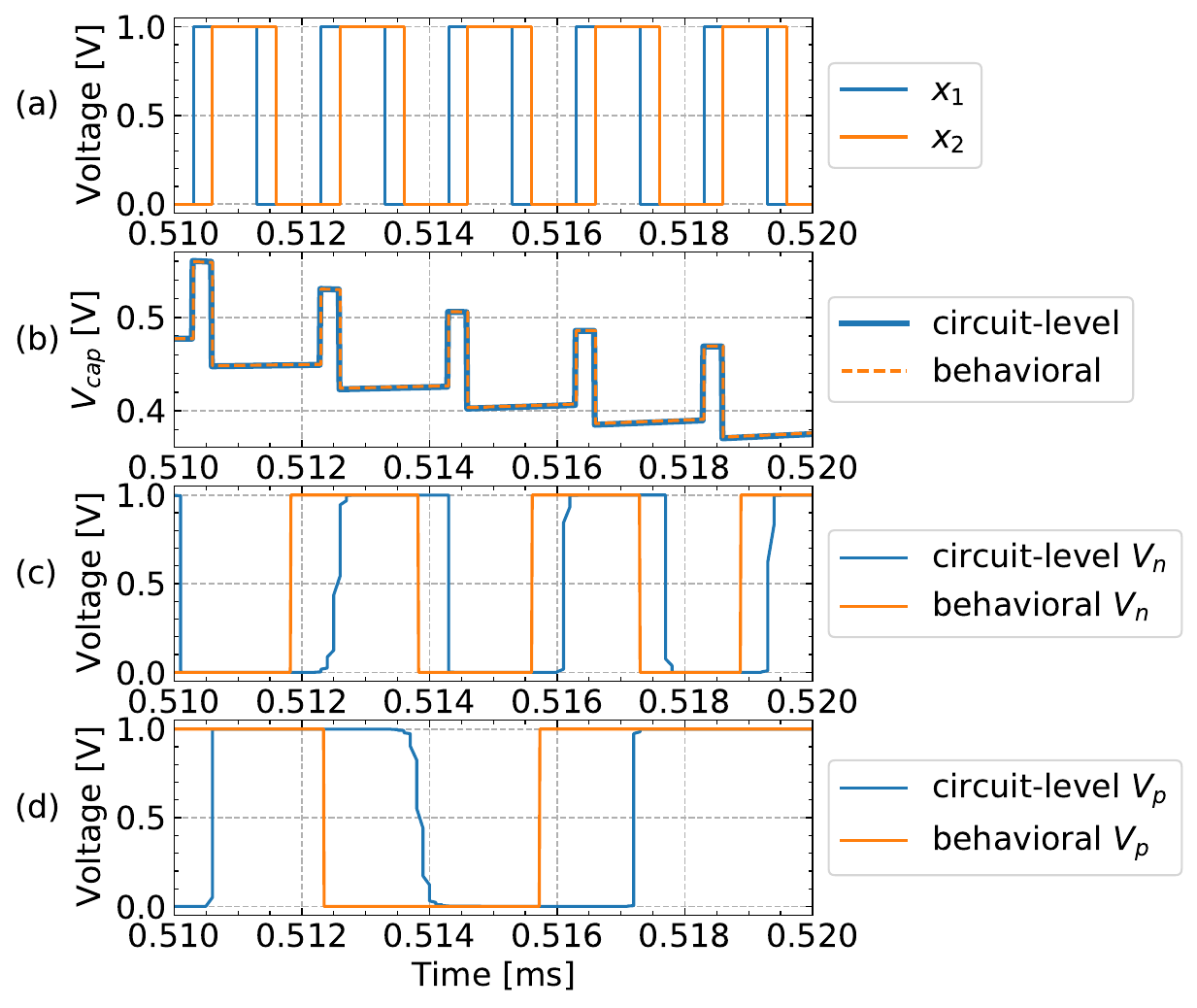}
\caption{\label{fig:all_compare} 
Comparison of $V_{cap}$ between the behavioral and the circuit-level simulations. The behavior of $V_{cap}$ is accurately reproduced by our model.
Comparison of VCO outputs between the behavioral model simulation and the HSPICE circuit-level simulation. The output frequencies of both $V_p$ and $V_n$ are accurately reproduced by the model.}
\end{figure}

Figure~\ref{fig:code} also shows how these 3 parts are applied in the behavioral model, where $m$ indicates the number of simulation steps. At each simulation step of 10\,ns, $V_{cap}$ is updated through Leakage and Integration blocks. The output frequency of $V_p$ and $V_n$ are determined by Fire block.
The feasibility of the constructed behavioral model is verified through comparisons with the circuit simulation results.
Figure~\ref{fig:behavior_test} shows the testbench circuit for comparison. 
The comparison between the behavioral model and the circuit simulation results of $V_{cap}$ along with the inputs $x_1$ and $x_2$ is depicted in Figures~\ref{fig:all_compare}(a) and (b). This result demonstrates that our model accurately reproduces the behavior of $V_{cap}$. 
Figures~\ref{fig:all_compare}(c) and (d) show the comparison of neuron outputs $V_{p}$ and $V_{n}$. The output frequencies agree well each other, although the phase of the oscillation does not align in this simulation since the initial phase of the oscillation is not well defined.

\section{Simulation and Results}

\subsection{System Structure}
\begin{figure}[tb]
\includegraphics[width=\linewidth]{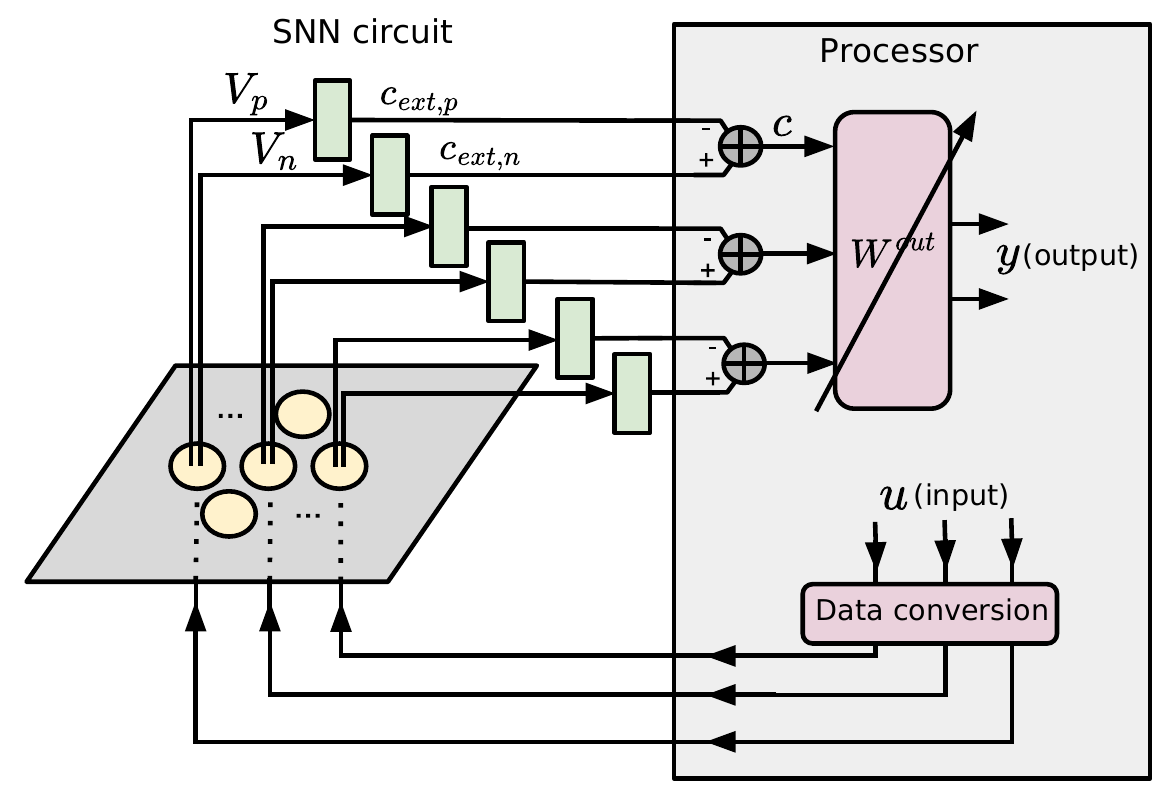}
\caption{\label{fig:chip_env} Whole simulation environment, which is composed of 100 neurons. The number of neurons is determined considering the feasible chip size when those are integrated into an IC chip. Neurons are supposed to be placed in an array on 2-dimensional plane. The number can be increased by enlarging the chip size. The input data, denoted as $u$, is transformed into a pulse train so that our neuron circuits described in Sect.~\ref{sec:circuit} can accept the input as a form of pulses. Every neuron fires pulses at a rate determined by its $V_{cap}$, which will be captured by the counter to be processed by an external processor.
}
\end{figure}

\begin{figure}[tb]
\includegraphics[width=\linewidth]{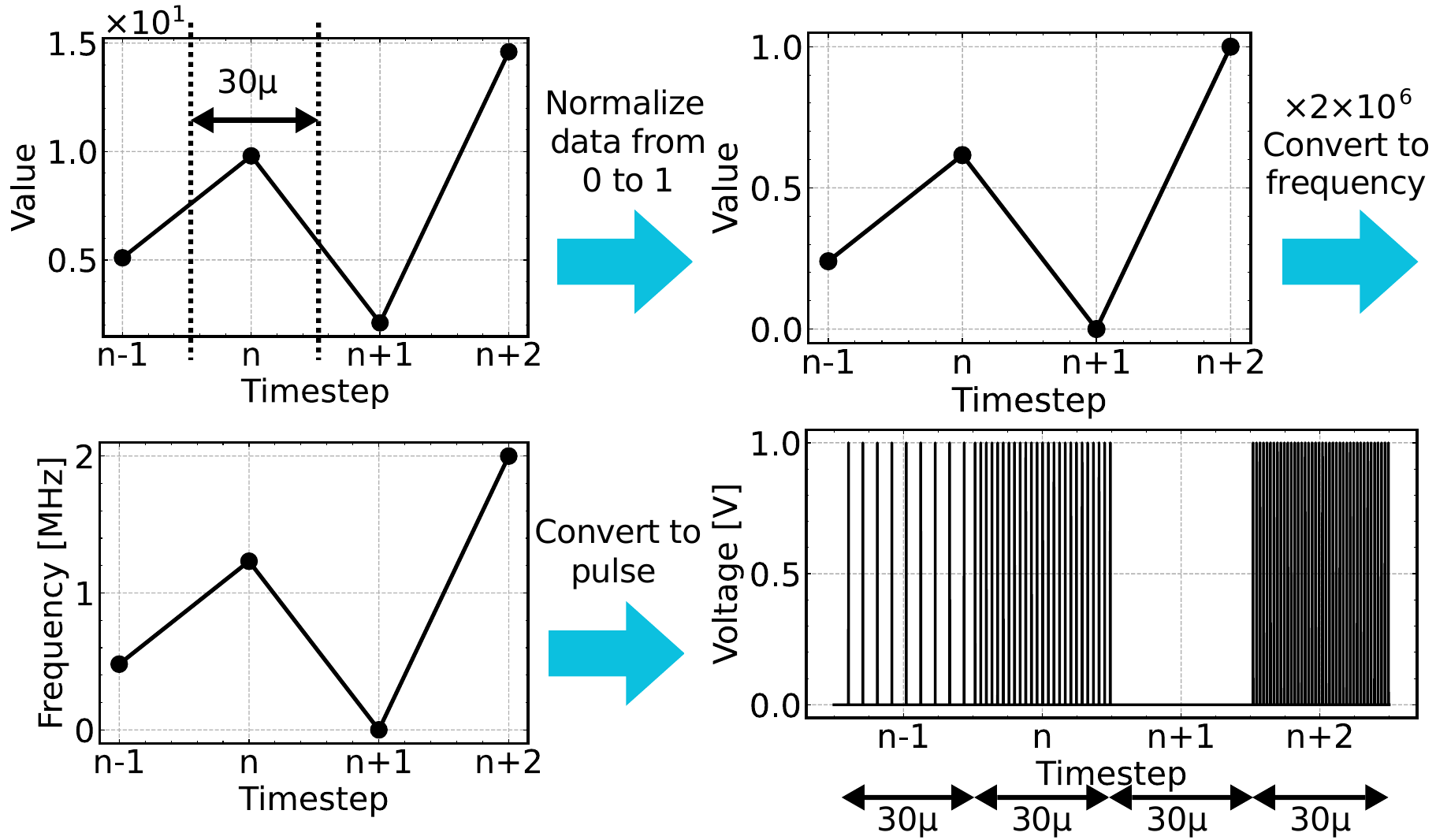}
\caption{\label{fig:data_convert} Procedure to convert data to pulse trains. The data are first normalized to the range of 0 to 1, and multiplied by $2\times 10^6$ to be converted to proper frequencies for our neurons. Then the pulse trains with corresponding frequencies are fed to the neuron circuits. The width of the pulses are fixed to 0.2\,ns in this experiment.}
\end{figure}

The diagram in Figure~\ref{fig:chip_env} depicts the entire architecture of the SNN system in our simulation.
The input data are in time-series form and discretized in time with a time step of 30\,$\mu$s, which are converted to pulses through the process outlined in Figure~\ref{fig:data_convert}. 

The final output of the SNN is given by
\begin{equation}
\bm{y} = W^{out}\cdot \bm{c} = W^{out}(\bm{c}_{ext,n} - \bm{c}_{ext,p}).
\end{equation}
Based on the error between $\bm{y}$ and a desired output $\bm{d}$, the output weights $W^{out}$ will be tuned during the learning process explained in the next subsection.

\subsection{Learning algorithm}
We utilize a linear regression~\cite{multivariate} for our learning algorithm to compute $W^{out}$
($\in \mathbb{R}^{N_{out}\times N}$), where $N$ and $N_{out}$ are the numbers of neurons and outputs, respectively.
Here we assume $\bm{d}(n)\,(\in \mathbb{R}^{N_{out}})$ as a teaching signal at $n$-th time step, and $\bm{x}(n)\,(\in \mathbb{R}^{N})$ as a representing vector of the reservoir at $n$-th time step. Our goal is to minimize the error $E_{LR}$ defined by
\begin{equation}
\label{Elr} E_{LR} = \frac{1}{2} \sum_{n=1}^T \lVert \bm{d}(n) - W^{out}\bm{x}(n) \rVert_2^2.
\end{equation}
In this study, we use counter values, denoted as $\bm{c}(n)$ to represent $\bm{x}(n)$. Let $X$ be a matrix that concatenates all $\bm{x}(n)$ up to time step $T$ ($X = [..., \bm{x}(n-1), \bm{x}(n), \bm{x}(n+1), ...] \in \mathbb{R}^{N\times T}$). Similarly, let $D$ be a matrix that concatenates all $\bm{d}(n)$ up to time step $T$ ($D = [..., \bm{d}(n-1), \bm{d}(n), \bm{d}(n+1), ...] \in \mathbb{R}^{N_{out}\times T}$). 
Then, the output layer $W^{out}$ can be calculated with~\cite{multivariate}
\begin{equation}
\label{eqx} W^{out} = DX^\mathsf{T}(XX^\mathsf{T})^{-1}.
\end{equation}

\subsection{Experiments}
\subsubsection{Delay task}
\begin{figure*}[tb]
    \includegraphics[width=0.8\linewidth]{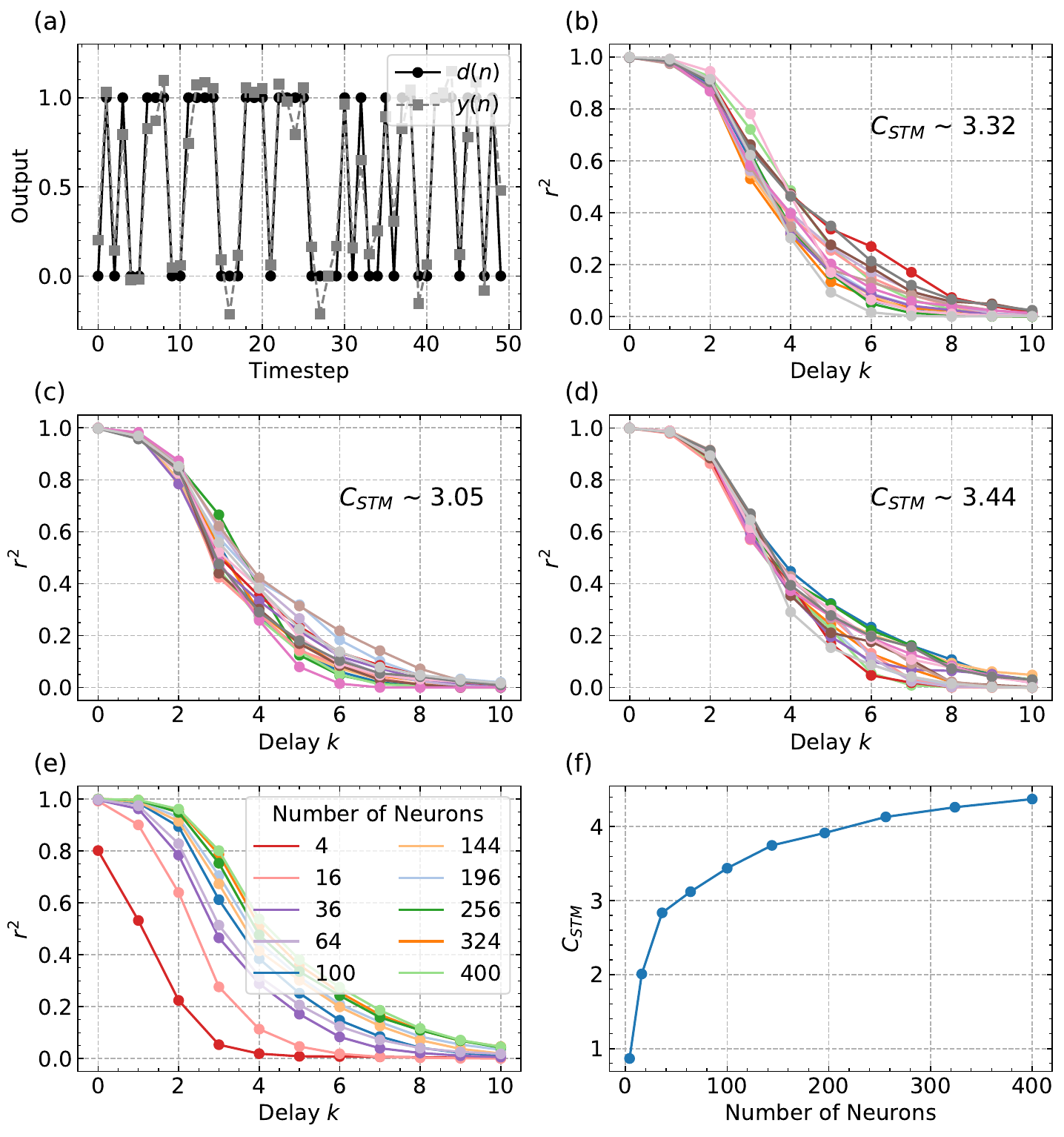}
    \caption{\label{fig:stm_all} 
    Experimental results of the delay task. 
    (a) The waveforms of the desired output $d(n)$ and the raw output data $y(n)$ in delay task with the delay $k=2$ with 10 $\times$ 10 neurons.
    (b) The score $r^2(k)$ of the delay task versus the delay $k$ for different weight assignments inside the reservoir with 10 $\times$ 10 neurons. The average memory capacity $C_{STM}$ is 3.32 and the standard deviation is 0.29.
    (c) The score $r^2(k)$ of the delay task versus the delay $k$ for the case only with the positive VCO. The average $C_{STM}$ is 3.05 and the standard deviation is 0.27. The $C_{STM}$ drops compared with the case with our proposed neuron circuit.
    (d) The score $r^2(k)$ of the delay task versus the delay $k$ considering the circuit parameter variation. The average $C_{STM}$ is 3.44 and the standard deviation is 0.25. The impact of the parameter variation due to the fabrication process is not significant compared with that of the weight assignment.
    (e) The score $r^2(k)$ of the delay task versus the delay $k$ depending on the number of neurons. As the number increases, $r^2(k)$ increases.
    (f) Relationship between the number of neurons and $C_{STM}$, which increases monotonically along with the number of neurons.}
    \end{figure*}

First, we perform a delay task to evaluate one of the key attributes of the reservoir, i.e., fading memory or short-term memory, as described in~\cite{jaeger2001short}. 
The input data $u(n)$ ($\in \mathbb{R}$) for this task is a sequence of random binary values, while the desired output $d(n)$ ($\in \mathbb{R}$) is given by
\begin{equation}
    d(n) = u(n-k) \qquad k=1,2,3...~,
\end{equation}
which is a simple $k$-step delayed version of the input sequence.
Supposing that $y(n)$ is the output of our SNN system, the correlation between $d(n)$ and $y(n)$ is defined as $r(k)$ as follows. 
\begin{equation}
    r(k)=
    \frac{\frac{1}{T} \sum_{n=1}^T (d(n)-\bar{d})(y(n)-\bar{y})}
    {\sqrt{\frac{1}{T} \sum_{n=1}^T (d(n)-\bar{d})^2} \sqrt{\frac{1}{T} \sum_{n=1}^T (y(n)-\bar{y})^2}}.
\end{equation}
Here, $\bar{d}=\frac{1}{T} \sum_{n=1}^T d(n)$ and $\bar{y}=\frac{1}{T} \sum_{n=1}^T y(n)$.
By summing up the square of $r(k)$, we can calculate the memory capacity $C_{STM}$ of the reservoir~\cite{jaeger2001echo} as given by
\begin{equation}
    C_{STM}=\sum_{k=1}^\infty r(k)^2.
\end{equation}
The experimental results are summarized in Figure~\ref{fig:stm_all}.

In our experiment, we used a total of 15000 random binary inputs for each delay $k$ with 1500 inputs used for washout, 10500 for training, and 3000 for testing. 
Since $r(k)^2$ becomes negligibly small for $k > 10$, we summed it up to $k=10$ to calculate $C_{STM}$.
Figure~\ref{fig:stm_all}(a) shows the waveforms of $d(n)$ and $y(n)$ with $k=2$ using 10 $\times$ 10 neurons for example, which shows that $y(n)$ properly predicts $d(n)$.

Since a Gaussian random value is assigned for each weight of the connection between neurons in the network, the dependence on the weight assignment is first investigated in Figure~\ref{fig:stm_all}(b), where 16 different weight assignments are compared.
The average memory capacity $C_{STM}$ is 3.32 and the standard deviation is 0.29 with 10 $\times$ 10 neurons. Though it varies depending on the weight assignments as expected, its impact is not so significant. 

To demonstrate the feasibility of the proposed neuron structure employing two VCOs with opposite sensitivity, we have tested the STM performance for the case only with the positive VCO as summarized in Figure~\ref{fig:stm_all}(c). The same 16 weight assignments as in (b) are compared, which shows the average $C_{STM}$ of 3.05 with the standard deviation of 0.27. For each weight assignment, the SNN only with the positive VCO results in lower STM score than the case with two VCOs, which proves the effectiveness of the proposed neuron structure.
In the following experiment, we fix the weight assignment with the case that achieves the best training score in the first experiment.

When the proposed SNN is actually implemented on a chip, the impact of the fabrication process variation or mismatch is inevitable. 
This leads to the fluctuation in neuron dynamics, which may cause the degradation in the memory capacity.
To verify the impact of this practical implementation issue, we mimic this fluctuation by randomly changing the dependence on $V_{cap}$ in the behavioral models of Leakage, Integration, and Fire in Figure~\ref{fig:code} for each neuron.
Figure~\ref{fig:stm_all}(d) plots $r(k)^2$ for 16 cases with random neuron fluctuation with a standard deviation of 30\,\%, which is a relatively pessimistic value considering the actual chip integration. The simulation results show that the average $C_{STM}$ is 3.44 and the standard deviation is 0.25. Compared with the case in Figure~\ref{fig:stm_all}(b), the variation is roughly the same, which shows that the impact of the process variation is not dominant considering the practical use.

Finally, Figures~\ref{fig:stm_all}(e) and (f) plot the $r(k)^2$ and $C_{STM}$ by changing the number of neurons. These plots use the mean values of $r^2(k)$ for 16 cases with random neuron fluctuation.
With more neurons the $C_{STM}$ improves, which demonstrates the scalability of the proposed architecture with larger-scale integration.

\subsubsection{XOR task}

    \begin{figure*}[tb]
    \includegraphics[width=0.8\linewidth]{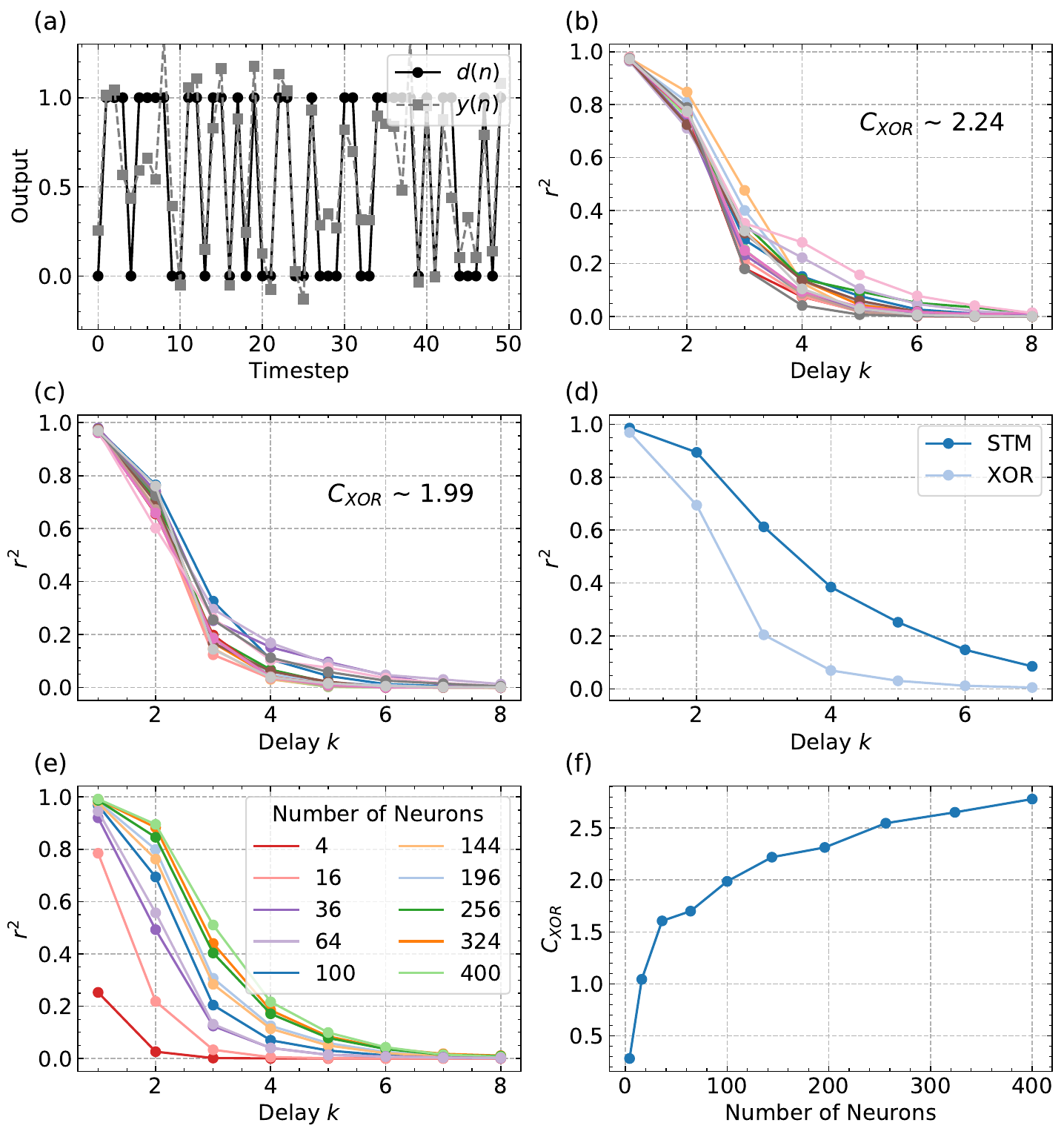}
    \caption{\label{fig:xor_all} 
    Experimental results of the XOR task. 
    (a) The waveforms of the teaching signal and the raw output data of the system in XOR task with the delay $k=2$ using 10 $\times$ 10 neurons. 
    (b) The score $r(k)^2$ of the XOR task versus the delay for different weight assignments inside the reservoir with 10 $\times$ 10 neurons. The average XOR capacity $C_{XOR}$ is 2.24 with the standard deviation of 0.18. 
    (c) The score $r(k)^2$ of the XOR task versus the delay considering the circuit parameter variation. The average $C_{XOR}$ is 1.99 and the standard deviation is 0.16. The impact of the parameter variation is more significant for the XOR task than the case for the delay task.
    (d) Comparison of the mean score of the delay task and the XOR task. The score $r^2(k)$ drops more steeply for the XOR task.
    (e) The score $r^2(k)$ of the XOR task versus the delay depending on the number of neurons. The score improves with more number of neurons.
   (f) Relationship between the number of neurons and $C_{XOR}$, which increases monotonically along with the number of neurons.}
    \end{figure*}

We next perform a temporal XOR task~\cite{jaeger2012long} to evaluate the high-dimensionality and nonlinearity of our SNN system.
The input is the same as that for the delay task, which is a random binary sequence. Here, the desired output $d(n)$ is defined as follows~\cite{toprasertpong2022reservoir}:
\begin{equation}
    d(n) = XOR(u(n), u(n-k)) \qquad k=1,2,3...~,
\end{equation}
which is a XOR of the current and the $k$-step delayed input bits.
$r(k)$ again represents the correlation between the output $y(n)$ and $d(n)$. Then we define the XOR capacity of the reservoir by summing up the square of this value as follows~\cite{toprasertpong2022reservoir}:
\begin{equation}
    C_{XOR}=\sum_{k=1}^\infty r(k)^2.
\end{equation}
We calculated $r(k)$ up to $k=7$, since $r(k)^2$ becomes negligible for $k>7$. 
Similarly to the delay task case, a total of 15000 random binary inputs including 1500 for washout, 10500 for training, and 3000 for testing, are used. 
Figure~\ref{fig:xor_all}(a) shows the waveforms of $d(n)$ and $y(n)$ with $k=2$ using 10 $\times$ 10 neurons for example, which shows that $y(n)$ properly predicts $d(n)$.

As verified in the STM test, the impact of the random weight and circuit fluctuation is verified. Figure~\ref{fig:xor_all}(b) summarizes the results for 16 different weight assignments, which shows the average XOR capacity $C_{XOR}$ of 2.24 with a standard deviation of 0.18 with 10 $\times$ 10 neurons. 
Figure~\ref{fig:xor_all}(c) shows the results for 16 cases with random neuron fluctuation with a standard deviation of 30\,\%, where the average $C_{XOR}$ drops to 1.99 with the standard deviation is 0.16. 
We found that the fluctuations of the design parameters affect more on the XOR capacity compared with the STM case.
Figure~\ref{fig:xor_all}(d) plots the mean of the STM results in Figure~\ref{fig:stm_all}(d) and the mean of the XOR results in Figure~\ref{fig:xor_all}(c) for comparison. As the delay $k$ increases, the XOR capacity decreases more steeply than the STM capacity. Since the XOR task requires the high-dimensionality and nonlinearity along with the fading memory, it is usual that the XOR capacity is lower than the STM capacity and that the XOR task is more sensitive to the delay $k$.

Figures~\ref{fig:stm_all}(e) and (f) plot the $r(k)^2$ and $C_{XOR}$ by changing the number of neurons using the mean values of $r^2(k)$ for 16 cases with random neuron fluctuation.
Similar to the STM case, the XOR performance improves with more neurons, 
which again demonstrates the scalability of the proposed system. We can clearly find that further improvement is expected with more neurons.

\renewcommand{\arraystretch}{1.5}
\begin{center}
\begin{table}[]
\caption{Comparison of capacity.}
\label{table:comparison}
\begin{tabular}{|l|c|c|c|c|} \hline
   Works & This work & \cite{toprasertpong2022reservoir} & \cite{nakane2021spin}$\ast$ & \cite{Bueno:17} \\ \hline
   Device & CMOS& FeFET & Spin waves & Laser \\ \hline
\#Nodes     & 400 & 600& 72& 330\\ \hline
\begin{tabular}{l}Virtual/Real$\star$\\Node\end{tabular} & Real & Virtual & Real & Virtual \\ \hline
Readout & \begin{tabular}{c}Counter\\circuits\end{tabular}& \begin{tabular}{c}Current\\measure\end{tabular}& \begin{tabular}{c}ADC\\circuits \end{tabular}&\begin{tabular}{c} Photo\\detector\end{tabular}\\ \hline
   $C_{STM}$ & 4.37& 2.35 & 16& 8.3 \\ \hline
   $C_{XOR}$ & 2.78 & 2.29 & 7.9& - \\ \hline
   \multicolumn{5}{l}{$\ast$ \cite{nakane2021spin} uses $d(n)=XOR(u(n-1),u(n-k))$ as XOR task.} \\
   \multicolumn{5}{l}{$\star$ ``Real'' means that the nodes are physically realized.} \\
\end{tabular}
\end{table}
\end{center}

Table~\ref{table:comparison} compares the STM and XOR capacities with other prior works.
The SNN with the proposed neuron exhibits better scores with smaller number of nodes compared with the FeFET device implementation~\cite{toprasertpong2022reservoir}, while the spin-wave and the laser devices~\cite{nakane2021spin,Bueno:17} demonstrate higher scores with smaller number of nodes. Here we note that the SNN presented in this paper assumes real nodes to compose the network 
considering the hardware-friendly network configuration with counter-based readout, while others use dedicated ADCs or analog quantities for readout. The $C_{STM}$ and $C_{XOR}$ scores for our SNN are achieved under more practical assumption of the actual network implementation. In addition, as demonstrated by the results in Figures~\ref{fig:stm_all}(f) and \ref{fig:xor_all}(f), we can expect further performance improvement by scaling up the number of nodes, which is feasible on the CMOS platform.

\begin{figure*}[tb]
\includegraphics[width=\linewidth]{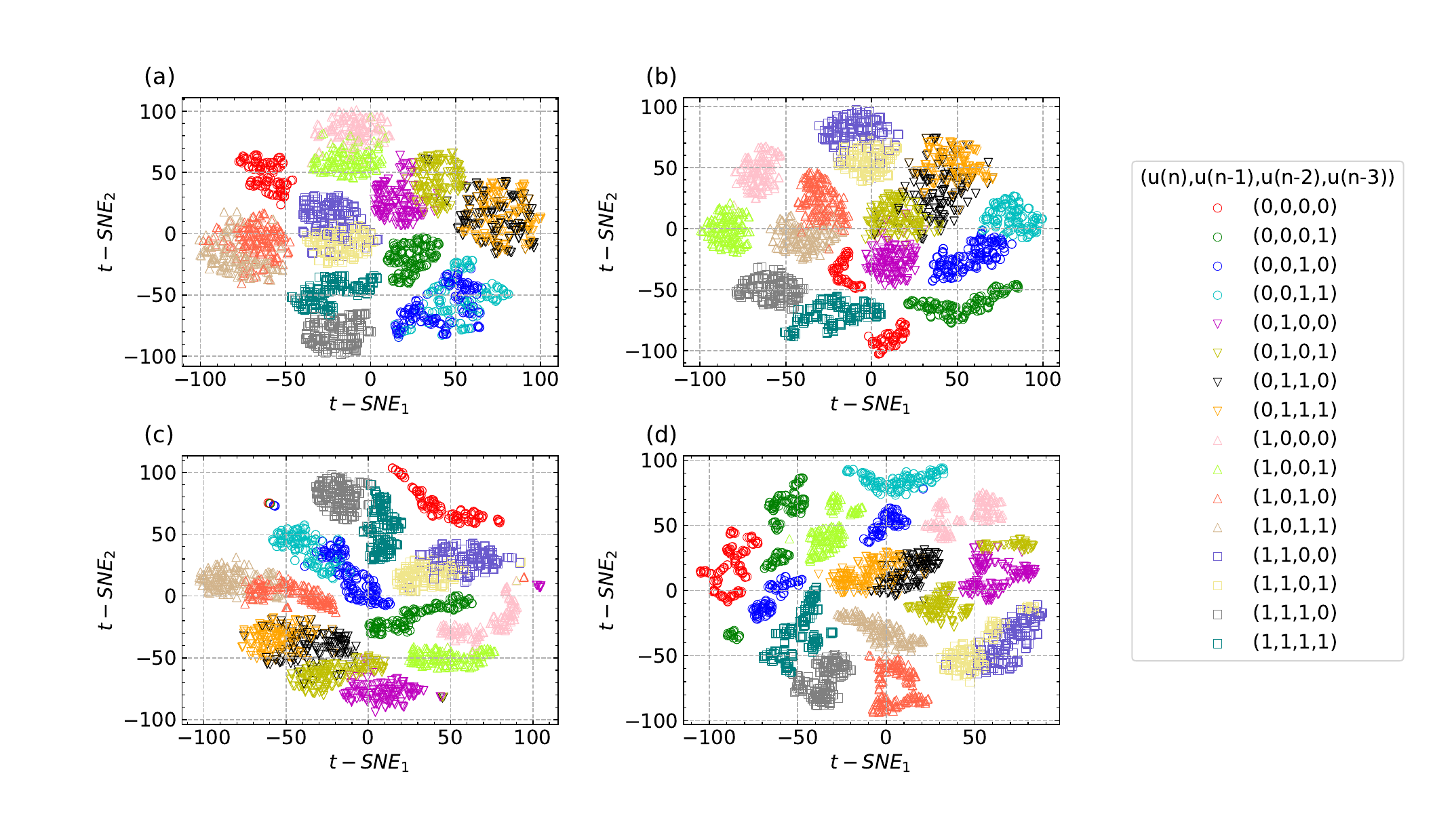}
\caption{\label{fig:tSNE_nodes} Counter values $\bm{c}$ is visualized with t-SNE with different colors coded by the inputs. The number of neurons are changed to see the improvement in the separation: (a) 49 nodes (b) 100 nodes (c) 196 nodes (d) 400 nodes.}
\end{figure*}

To gain better understanding of the reservoir behavior, it is useful to visualize a high-dimensional counter output generated by the neural network, which can be achieved by converting the high-dimensional output to a 2-dimensional format. 
Figure~\ref{fig:tSNE_nodes} depicts the reservoir output $\bm{c}(n)$ of the XOR task using t-SNE visualization~\cite{van2008tsne} for different number of neurons. The color-coding distinguishes different input sequences for the previous three time steps, i.e., ($u(n), u(n-1), u(n-2), u(n-3)$). 
By taking a look at Figure~\ref{fig:tSNE_nodes}(b), for the case with 100 neurons, 
the orange and black points are closely located on the 2-dimensional plane. 
This mixture results in a decreased score when $k=3$. Black and purple points have the same inputs except for $u(n-2)$ and locate relatively apart from each other, which explains that the score when $k=2$ is higher than that when $k=3$.
As the number of nodes increases, the regions with overlap by different colors decrease, which implies that the SNN distinguishes the different input sequences more efficiently with more neurons. This trend is observed also in STM and XOR capacities as shown in Figures~\ref{fig:stm_all}(f) and \ref{fig:xor_all}(f). These results prove that the implemented learning algorithm properly works to train the weight of the output layer.

\subsubsection{Spoken Digit Recognition}
\begin{figure*}[tb]
\includegraphics[width=\linewidth]{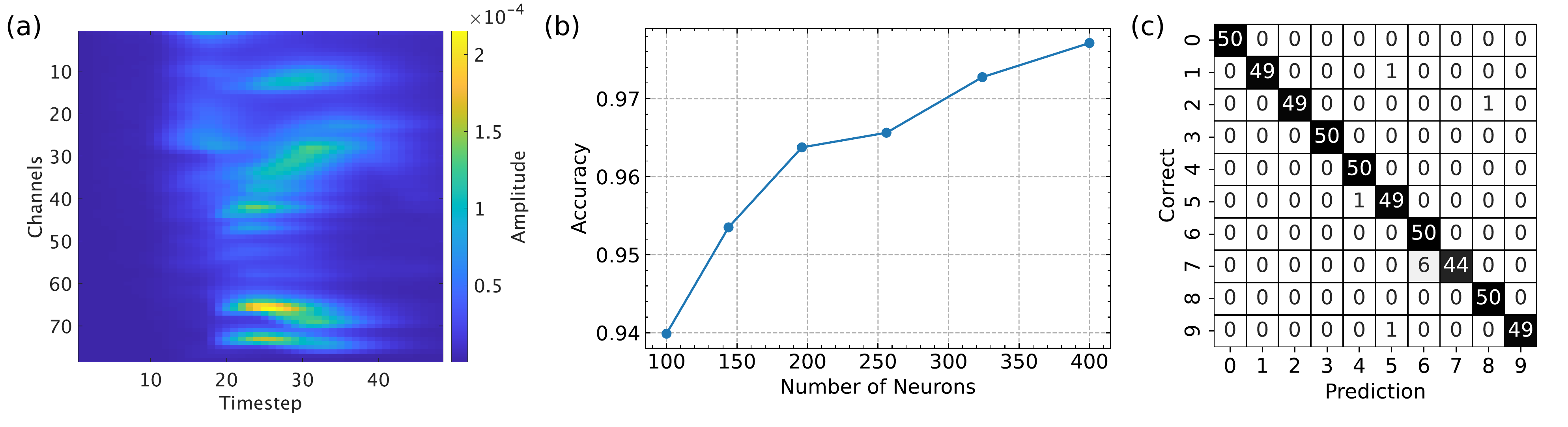}
\caption{\label{fig:spoken_digit_result} 
(a) Example of a spoken digit data after processed with Lyon's cochlear model. 
(b) Relationship between the number of neurons and the accuracy, which increases monotonically along with the number of neurons.
(c) Confusion matrix in spoken digit recognition task using the proposed SNN circuit with 400 neurons with no fluctuation. The horizontal axis shows the predicted label and the vertical axis shows the correct label. The number of data categorized to each bin is summarized.
}\end{figure*}

As the dynamics of the proposed neural network such as high-dimensionality, nonlinearity, and fading memory have been verified with both STM and XOR tasks,
next we applied the system to a more meaningful recognition task, spoken digit recognition~\cite{verstraeten2005isolated}.
We utilized TI46-Word dataset~\cite{ti46}, which includes speech data for 46 different words. Specifically, we used 500 instances of female speech data for 0-9 digits. 
As a preprocessing, we employed Lyon's cochlear model~\cite{lyon1982computational}, which replicates the human ear system, before feeding the data to the SNN using the LyonPassiveEar function in Auditory Toolbox~\cite{slaney1998auditory}. 
We set the sampling rate to 12.5kHz and set the decimation factor to make the time step of each data 48. Other parameters are set to the default values. This results in a matrix of $78\times 48$, whose example is shown in Figure~\ref{fig:spoken_digit_result}(a). We then remove the first 10 and the last 8 silent parts, resulting in a length of 30 for each instance.

Then the preprocessed data are partitioned into 5 segments and subjected to cross-validation, 4 of which are reserved for training and the rest 1 is used for testing. 
To recognize 10 spoken digits, we prepare 10 output nodes, each corresponds to a predicted label. During the training, the output node for the correct label is trained to be 1, while the rest 9 nodes to be $-1$.

To quantify the performance of the reservoir, the recognition accuracy of our SNN system is first compared with a simple linear regression without a reservoir, which achieved 86.4\,\%.
Similarly to the case of STM and XOR tasks in the previous subsection, we first tried 16 random weight assignments. Among the 16 cases, we used the weight assignment that achieves the best training accuracy for the following test. In this experiment, the recognition accuracy is up to 95.2\,\%, which is higher than the case with a simple linear regression. This properly demonstrates the effectiveness of the SNN. 

Then, considering the practical chip implementation, the impact of the process variation is considered in the same way as in the STM and XOR tasks. Again, 30\,\% performance fluctuation is assumed. As a result, mean accuracy of 94.0\,\% with the standard deviation of 0.8 percentage point is achieved. Though a slight decrease in accuracy is observed, still we can achieve higher performance than a simple linear regression case.

As in the cases of the STM and XOR tests, the recognition accuracy improves along with the number of neurons.
Figure~\ref{fig:spoken_digit_result}(b) plots the accuracy by changing the number of neurons using the mean accuracy values for 16 cases with random neuron fluctuation.
When 196 neurons are used, the accuracy is 96.4\,\% with a standard deviation of 0.8 percentage point, while it reaches 97.7\,\% with a standard deviation of 0.5 percentage point with 400 neurons.
The confusion matrix for the case with 20 $\times$ 20 neurons is shown in Figure~\ref{fig:spoken_digit_result}(c) for the reference. 
The accuracy improves with more neurons, which also confirms the scalability of the proposed system.

Table~\ref{table:ti46_comparison} summarizes the comparison of the recognition accuracy with some prior works, which shows that our SNN system achieves a comparable recognition performance considering the hardware-friendly network configuration with counter-based readout on CMOS platform. Note that our SNN implementation uses 20×20 real nodes while others assume virtual nodes for this verification.

\renewcommand{\arraystretch}{1.5}
\begin{table*}[tb]
    \centering
    \caption{Comparison of accuracy in spoken digit recognition task.}
    \label{table:ti46_comparison}
    \begin{tabular}{|c|c|c|c|c|c|c|} \hline
       Works & This work & \cite{appeltant2011information}& \cite{paquot2012optoelectronic}& \cite{duport2012all} & \cite{moon2019temporal} & \cite{martinenghi2012photonic}\\ \hline 
    Device & CMOS & Circuit $\ast$ & Optolelectronic & Optic & Memristor & Photonic \\ \hline
    \#Nodes     & 400& 400 & 200& 200& 400& 150\\ \hline
    Virtual/Real Node & Real & Virtual & Virtual & Virtual & Virtual & Virtual \\ \hline
    Readout& Counter& ADC& Photodiode& Photodiode& Conductance(Current)& ADC\\ \hline
    Accuracy & 97.7\%& 99.8\%& 99.6\%&97.0\%&99.2\%&99.4\%\\ \hline
    \multicolumn{7}{l}{$\ast$ Using discrete devices \cite{appeltant2011informationsupp}} \\
    \end{tabular}
\end{table*}

\section{Discussion}
In this paper, we revealed the potential advantages of the proposed CMOS-based time-domain spiking neuron. Especially the scalability of the CMOS-based implementation is well demonstrated with multiple experimental results. Since the proposed neuron is implemented with a small area, e.g., just a 30\,$\mu$m $\times$ 70\,$\mu$m area for a single neuron with relatively mature 65\,nm CMOS process in our previous work~\cite{chen2023cmos}, it is straightforward to integrate a few hundred neurons onto one chip with a practical size. As our network configuration also takes into account the connection between neurons considering the 2-dimensional on-chip integration, the wiring complexity will not explode with the larger number of neurons. 

Another advantage we would emphasize is that the proposed neural network system uses a hardware-friendly counter-based readout to capture the internal state of each neuron, where  neurons and readouts are connected with each other on a one-to-one basis. The feature facilitates the use of many real nodes with the practical, small-area, and low-power-consumption readout circuits, unlike the prior works (Table II) that use virtual nodes with complicated, large-area, and high-power-consumption readout ADC circuits. 

The primary goal of this study is to demonstrate the basic capability of the proposed system, using the simple physical structure and computing procedure. Hence, to improve the computational performance, we can feasibly explore additional schemes, e.g., the system size, fine-tuning, a reservoir input matrix, and the time step duration, by leveraging the practical advantages of CMOS-based systems, as discussed below.

As demonstrated by the experimental results in Figure~\ref{fig:spoken_digit_result}(b), we expect performance improvement by scaling up the number of nodes. This guideline is extremely favorable, since the CMOS implementation is one of the most suitable platforms for integrating massive number of elements within a single chip. 

Whereas this study uses the fixed parameters in dynamics, such as the oscillation frequency and the leakage time constants, we can fine-tune those parameters by circuit, i.e., we can feasibly tune the neuron dynamics by design. This can offer a major advantage over other reservoir systems on a chip.

In this study, the sequential input is fed into all the neurons and the reasonably high performance is achieved. Nonetheless, this input scheme possibly leads to lower computational capability of the system, since similar output waveforms in neighboring neurons are presumably generated due to significant influence of input on each neuron dynamics and consequently the effective dimensionality of the reservoir output vector is probably not so high. As frequently used in ESNs ~\cite{jaeger2001echo, jaeger2012long}, a part of neurons can be selected as input neurons through a reservoir input matrix that defines the connections between the one-dimensional input part and the neurons. We expect that a sparse matrix can result in diverse output waveforms and higher computational performance is consequently achieved. Moreover, since the input is crucial for nonlinear reservoir dynamics, another magnitude of the input and/or another time step duration can find higher computational performance. We can explore those input schemes by simply changing the input procedure using the same hardware chip including peripheral circuits, which is certainly one of the advantages of CMOS-based systems.

\section{Conclusion}
This paper proposed a CMOS-based time-domain analog spiking neuron employing two VCOs with opposite sensitivities to the internal control voltage. By utilizing the proposed neuron circuit, we can compose the SNN with a counter-based readout circuit, which is friendly to the actual hardware implementation.
To quantify the performance of the reservoir based on our SNN, we built the behavioral models of the proposed neuron circuits for the system-level simulations. In the behavioral simulation considering the restrictions of the on-chip implementation, the feasibility of the proposed SNN system is verified with STM and XOR tasks. Spoken digit recognition task was also conducted to evaluate the performance of our hardware-friendly implementation of the SNN, which resulted in 97.7\,\% recognition accuracy with 20 $\times$ 20 neurons.

Every spiking neuron in our network-based system is accessible to detect its dynamical state that can originate from fading and coupled oscillation phenomena frequently seen in various physical dynamical systems. This prominent feature can contribute to deep understanding of the relationship between physical dynamics and computational capability as well as to the establishment of the system design for high computational performance in practical applications. Therefore, our system offers a very useful platform to advance the research field of physical RC toward machine-learning electronics with high performance and low power consumption.

\section*{Acknowledgements}
This work was supported by the Japan Science and Technology Agency (JST) CREST (Grant No. JPMJCR19K2).

\bibliography{apssamp}% Produces the bibliography via BibTeX.

\end{document}